\def\year{2020}\relax
\documentclass[letterpaper]{article} 
\usepackage{aaai20}  
\usepackage{times}  
\usepackage{helvet} 
\usepackage{courier}  
\usepackage[hyphens]{url}  
\usepackage{graphicx} 
\usepackage{graphicx}  
\usepackage{comment} 
\usepackage{subcaption,amsmath,bbm,nameref, multirow}

\usepackage[noend]{algpseudocode}
\usepackage{times}
\usepackage{fancyhdr,graphicx,amsmath,amssymb}
\usepackage[]{algorithm2e}
\usepackage[table]{xcolor}
\newcommand\norm[1]{\left\lVert#1\right\rVert}

\title{Building Calibrated Deep Models via Uncertainty Matching with Auxiliary Interval Predictors}
\author{Jayaraman J. Thiagarajan$^{\dagger}$\thanks{This  work  was  performed  under  the  auspices  of  the  U.S.  Departmentof Energy by Lawrence Livermore National Laboratory under Contract DE-AC52-07NA27344.}, Bindya Venkatesh$^{\ddagger}$, Prasanna Sattigeri$^{+}$, \\ \Large \bf Peer-Timo Bremer$^{\dagger}$
\\
$^{\dagger}$Lawrence Livermore National Laboratory, $^{\ddagger}$Arizona State University, $^{+}$IBM Research AI
\\
jjayaram@llnl.gov, bindya.venkatesh@asu.edu, psattig@us.ibm.com, bremer5@llnl.gov
}
 
\begin{document}

\maketitle

\begin{abstract}
With rapid adoption of deep learning in critical applications, the question of when and how much to trust these models often arises, which drives the need to quantify the inherent uncertainties. While identifying all sources that account for the stochasticity of models is challenging, it is common to augment predictions with confidence intervals to convey the expected variations in a model's behavior. We require prediction intervals to be well-calibrated, reflect the true uncertainties, and to be sharp. However, existing techniques for obtaining prediction intervals are known to produce unsatisfactory results in at least one of these criteria. To address this challenge, we develop a novel approach for building calibrated estimators. More specifically, we use separate models for prediction and interval estimation, and pose a bi-level optimization problem that allows the former to leverage estimates from the latter through an \textit{uncertainty matching} strategy. Using experiments in regression, time-series forecasting, and object localization, we show that our approach achieves significant improvements over existing uncertainty quantification methods, both in terms of model fidelity and calibration error.
\end{abstract}

\section{Introduction}
Data-driven learning methods, in particular deep learning, have led to incredible advances in a wide variety of commercial AI applications, and are rapidly being adopted in high impact applications such as healthcare, autonomous driving, and scientific discovery. Consequently, it has become critical to enable deep models to reliably assess their confidence in scenarios different from the training regime and to verify that predictions arise from generalizable patterns rather than artifacts in the training data. In practice, a variety of sources including data sampling errors, model approximations and the inherent data uncertainties can contribute to stochasticity in deep models~\cite{gal2016uncertainty}. Hence, capturing these uncertainties when making predictions can shed light onto when and how much to trust the results~\cite{healthuqleveraging,uqbioopportunities,cvuncertainties,levasseur2017uncertaintiesphysics}. Furthermore, accurate uncertainty estimates are essential for solving a variety of practical machine learning challenges including knowledge transfer~\cite{wen2019bayesian}, active learning~\cite{wang2016cost}, and anomaly detection~\cite{leibig2017leveraging}.

In recent years, several classes of uncertainty estimation techniques have been proposed~\cite{gal2016dropout,gal2017concrete,lakshminarayanan2017ensemble,ghahramani2015probabilistic}.
An inherent limitation of these uncertainty quantification (UQ) techniques is the lack of suitable validation strategies. A typical workaround is to evaluate prediction intervals (PIs) via their calibration, without distinguishing the uncertainty sources~\cite{heskes1997practical}: A PI is well calibrated if the likelihood of true target falling in the interval is consistent with the confidence level of the interval. In the context of calibration, most existing UQ methods in deep learning are not inherently calibrated~\cite{kuleshov2018accurate}.

In this paper, we argue that producing well-calibrated prediction intervals requires the identification of informative and uninformative uncertainties. When a model is equally confident in all parts of the data space, the uncertainties are considered uninformative. In practice, it is typical to use an additional recalibration dataset to refine the uncertainty estimates. In contrast, we propose to effectively leverage the informative uncertainties from a model through the use of an auxiliary interval estimator that is exclusively optimized for the calibration objective, and a novel \textit{uncertainty matching} process. More specifically, we use a primary network for producing mean estimates with prediction uncertainties and an auxiliary network for obtaining calibrated PIs, and formulate a bi-level optimization approach for jointly training the two networks. To this end, we develop two \textit{uncertainty matching} strategies to leverage the intervals during the mean estimation process, namely \textit{Sigma Fit} and \textit{IQR Fit}. By incorporating the calibration step into the training process, our method is able to produce predictive models with significantly improved accuracy as well as better calibration.

\subsection{Contributions:}
\begin{itemize}
	\item A novel framework to produce reliable prediction intervals in continuous-valued regression, that are sharp and well calibrated for any required confidence level;
	\item An approach to estimate means and prediction intervals as a bi-level optimization and an alternating optimization strategy for model inference;
	\item A new regularization for estimating means along with prediction uncertainties based on \textit{uncertainty matching};
	\item Two different \textit{uncertainty matching} strategies, namely \textit{Sigma Fit} and \textit{IQR Fit}; and
	\item A rigorous evaluation on different use-cases and model architectures including regression with FCNs, time-series forecasting with LSTMs and object localization with CNNs, all producing better generalization performance and calibration error compared to the state of the art.
\end{itemize}

\section{Related Work}
Broadly, there are two important sources of prediction uncertainties in deep models: epistemic uncertainty, also known as model uncertainty that can be eliminated given enough training data, and aleatoric uncertainty, which depends on stochasticity inherent to the data. A variety of estimators have been proposed in the literature for measuring these uncertainties, most often with classification models. For example, Bayesian neural nets~\cite{blundell2015weight}, Monte-Carlo dropout~\cite{gal2016dropout}, concrete dropout~\cite{gal2017concrete} and  ensemble techniques~\cite{lakshminarayanan2017ensemble} are commonly utilized to estimate the epistemic uncertainties. On the other hand, Tagasovska \textit{et al.} recently proposed a conditional quantile based estimator for measuring aleatoric uncertainties~\cite{tagasovska2018frequentist}.

Due to the lack of suitable evaluation mechanisms for validating the quality of these estimates, it is common to utilize calibration as a quality metric. Interestingly, it has been reported in several studies that these estimators are not inherently calibrated~\cite{kuleshov2018accurate}. Consequently, a large class of techniques that are aimed at calibrating pre-trained classifiers has been developed~\cite{guo2017calibration,seo2019singleshot,nixon2019measuring,kuleshov2015calibrated}. In comparison, calibration for regression tasks has received limited attention~\cite{gneiting2007strictly,kuleshov2018accurate,levi2019evaluating}. For example, a two-stage recalibration technique was proposed in~\cite{kuleshov2018accurate} which re-calibrates the PI estimates from the classifier model. This approach requires sufficient i.i.d.\ data to perform recalibration and can arbitrarily alter the true uncertainties to achieve the calibration objective~\cite{levi2019evaluating}. Another important limitation of recalibration methods is that they do not distinguish between informative and uninformative uncertainties. In order to circumvent this challenge, recently, Levi \textit{et al.} proposed to match the heteroscedastic variances to the residuals from the predictions~\cite{levi2019evaluating}, through a strategy similar to temperature scaling~\cite{guo2017calibration}. In contrast, our approach attempts to utilize the informative uncertainties for producing calibrated intervals, through an auxiliary interval estimator model and a novel \textit{uncertainty matching} strategy.
\begin{figure}[t]
	\centering
	\includegraphics[width=0.95\linewidth]{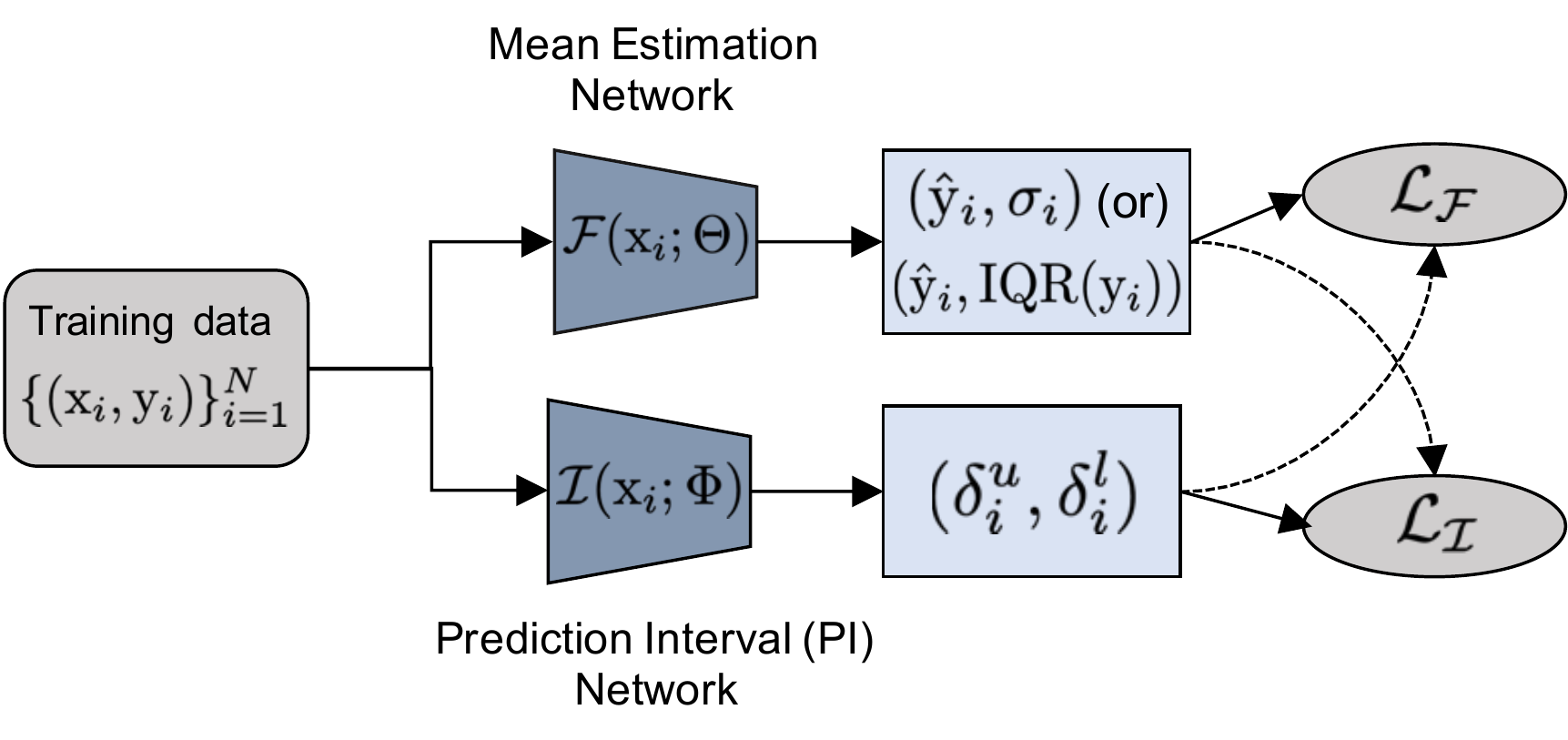}
	\caption{An illustration of our approach to produce PI, for any specified confidence $\alpha$, and perform PI-regularized mean estimation via a bi-level optimization technique.}
	\label{fig:BD}
\end{figure}

\section{Proposed Approach}
In this section, we describe the proposed approach (see Figure~\ref{fig:BD}) for building deep predictive models, which employs a novel \textit{uncertainty matching} strategy to produce estimators that are both accurate and well-calibrated.

\noindent \textbf{Notations.} Denoting the input data by $\mathrm{x} \subset \mathcal{X} \in \mathbb{R}^d$, our goal is to build a predictive model $\mathcal{F}: \mathrm{x} \mapsto \mathrm{y}$, where $\mathrm{y} \subset \mathcal{Y} \in \mathbb{R}$ and an interval estimator $\mathcal{I}: \mathrm{x} \mapsto \mathbb{R}^2$. Here, $\mathcal{X}$ and $\mathcal{Y}$ indicate spaces of inputs and outputs respectively. Note, the model $\mathcal{I}$ produces $\delta^u$ and $\delta^l$, which denote widths of the interval in the positive/negative direction respectively, thus producing a total interval width $w = \delta^u + \delta^l$. The models $\mathcal{F}$ and $\mathcal{I}$ are described using parameters $\Theta$ and $\Phi$ respectively. A labeled dataset of $N$ samples is comprised of tuples $\{(\mathrm{x}_i, \mathrm{y}_i)\}, i = 1, \cdots, N$. We denote all estimated quantities with a hat symbol, e.g. $\hat{\mathrm{y}} = \mathcal{F}(\mathrm{x})$. Though our focus is on predicting continuous-valued outputs, the proposed techniques can be easily generalized to classification problems.

\subsection{Formulation}
In this paper, we hypothesize that integrating the task of calibration into the predictive modeling process can provide useful regularization to its optimization, thus producing higher fidelity models that are also well-calibrated. To this end, our approach employs two separate models -- one for predicting the target and the other for estimating the prediction intervals, and pose a bi-level optimization formulation that allows the mean estimator to identify prediction uncertainties that are the most informative for matching the intervals from the PI estimator. Let $\mathcal{F}(\mathrm{x}; \Theta)$ denote the mean estimator and $\mathcal{I}(\mathrm{x}; \Phi^*)$denote the PI estimator parameterized by $\Theta$ and $\Phi$ respectively. The overall objective is
\begin{align}
\nonumber &\min_{\Theta} \mathrm{L}_{\mathcal{F}}\left(\Theta; \mathrm{x}, \mathrm{y}, \mathcal{I}(\mathrm{x}; \Phi^*)\right), \\
&\text{s.t.} \quad \Phi^* = \arg \min_{\Phi} \mathrm{L}_{\mathcal{I}}(\Phi; \mathrm{x}, \mathcal{F}(\mathrm{x}; \Theta)).
\label{eqn:obj}
\end{align}
Here, the first term represents an improved mean estimator that takes the intervals into account for identifying informative uncertainties, and the second term exclusively optimizes for calibration, based on the current mean estimator. Intuitively, the auxiliary task of obtaining prediction intervals, i.e., minimizing $\mathrm{L}_{\mathcal{I}}$, is used to regularize the primary task of fitting a predictive model by constraining its uncertainty estimates to match the estimated intervals. We refer to this process as \textit{uncertainty matching}. Though this can be implemented with any existing uncertainty estimation technique, we use heteroscedastic neural networks (HNN) and conditional quantile estimators in our setup.

\noindent \textbf{PI Estimator Design.} While the mean estimate is a random variable, an interval estimate is a random interval. For an output variable $\mathrm{y}$, let $\mathrm{y}^l$ and $\mathrm{y}^u$ denote its lower and upper bounds. While an interval does or does not contain a certain value, a random interval has a certain probability of containing a value. Suppose that $p(\mathrm{y}^l \leq \mathrm{y} \leq \mathrm{y}^u)= \alpha, \text{ where } \alpha \in [0, 1],$ then the random interval $[\mathrm{y}^l, \mathrm{y}^u]$ is referred as a $100\times \alpha\%$ interval of $\mathrm{y}$. When this holds, we expect that when we sample a large number of $\mathrm{x_i}$ from a validation set and predict $\mathcal{F}(\mathrm{x_i})$ and $\mathcal{I}(\mathrm{x_i})$, then $\mathrm{y_i}$ will be contained in $\mathcal{I}(\mathrm{x_i})$ $100\times \alpha\%$ of the trials.

The PI estimator network is designed to produce intervals that are well-calibrated and can reflect the underlying prediction uncertainties. In contrast to existing approaches which attempt to produce uncertainties that can be calibrated to any $\alpha$ by applying appropriate scaling, our method operates with a specific $\alpha$ and implements the calibration task as a differentiable function. As expressed in Eq. (\ref{eqn:obj}), the loss $\mathrm{L}_{\mathcal{I}}$ is constructed based on the current state of $\Theta$ for the mean estimator network $\mathcal{F}$. We use the following term to optimize for the quality of calibration, which measures the empirical probability of the true target lying in the estimated intervals:
\small
\begin{equation} \label{eqn:emce}
\mathrm{L}_{emce} =  \left|\alpha - \frac{1}{N}{ \displaystyle \sum_{i=1}^{N}\mathbbm{1}\left[(\hat{\mathrm{y}}_{i}-\delta_i^l) \leq \mathrm{y}_i \leq (\hat{\mathrm{y}}_{i}+\delta_i^u)\right]} \right|,
\end{equation}\normalsize where $\mathbbm{1}$ is the indicator function and $\hat{\mathrm{y}}_i = \mathcal{F}(\mathrm{x}_i; \Theta)$ are the mean predictions. We implement the indicator function as $\texttt{Sigmoid}(\eta \times (\mathrm{y}_i - \mathrm{y}_i^l)(\mathrm{y}_i^u - \mathrm{y}_i))$, where $\hat{\mathrm{y}}_i^l = \hat{\mathrm{y}}_i - \delta_i^l$, $\hat{\mathrm{y}}_i^u = \hat{\mathrm{y}}_i + \delta_i^u$, and $\eta$ is a large scaling factor. This empirical objective does not require any distributional assumption on data.  Next, in order to ensure that the intervals reflect the underlying uncertainty, the widths $w_i$ are matched to the residual $r$ from $\mathcal{F}$.
\begin{equation} \label{eqn:noise}
 \mathrm{L}_{noise}= \sum_{i=1}^N \norm{ \biggl(0.5 * w_i - |\mathrm{r}_i| \biggr)}_1,
\end{equation}where $\|.\|_1$ denotes the $\ell_1$ norm, $w_i = \delta^u_i + \delta^l_i$ and the residual $\mathrm{r}_i = \mathrm{y}_i - \mathcal{F}(\mathrm{x}_i; \Theta)$. Finally, the widths of estimated intervals are regularized to avoid non-trivial solutions of arbitrarily large intervals.
\begin{equation} \label{eqn:sharp}
 \mathrm{L}_{sharp} =  \sum_{i=1}^N \norm{\hat{\mathrm{y}}_i^u-\mathrm{y}_i}_1 +\norm{\mathrm{y}_i-\hat{\mathrm{y}}_i^l}_1.
\end{equation}Hence the overall objective can be written as:
\begin{align} \label{eqn:conf esti}
\nonumber \Phi^* &= \arg \min_{\Phi} \mathrm{L}_{\mathcal{I}} \\
&= \arg \min_{\Phi} \mathrm{L}_{emce} + \beta_n \mathrm{L}_{noise} + \beta_s \mathrm{L}_{sharp},
\end{align}where $\beta_n$, $\beta_s$ are penalty terms.

\RestyleAlgo{boxruled}
\begin{algorithm}[t]

	\KwIn{Labeled data $\{(\mathrm{x}_i, \mathrm{y}_i)\}_{i=1}^N$, Desired calibration level $\alpha$, Number of epochs $n_{m}$ and $n_c$.}
	\KwOut{Trained mean and interval estimators $\mathcal{F}$ and $\mathcal{I}$}
	\textbf{Initialization}:Randomly initialize parameters $\Theta^*,\Phi^*$\;
	\While{not converged}{
		\For{$n_{m}$ epochs}{
			Compute intervals $\delta_i^u, \delta_i^l = \mathcal{I}(\mathrm{x}_i; \Phi^*)$ \;
			Compute loss function $\mathrm{L}_{\mathcal{F}}$ using Eq. (\ref{bnn matching}) for \textit{Sigma Fit} or Eq. (\ref{IQR matching}) for \textit{IQR Fit} \;
			Update $\Theta^* = \arg \min_{\Theta} \mathrm{L}_{\mathcal{F}} $ \;
		}\For{$n_{c}$ epochs}{
			Obtain predictions $\hat{\mathrm{y}}_i = \mathcal{F}(\mathrm{x}_i; \Theta^*)$ \;
			Compute loss function $\mathrm{L}_{\mathcal{I}}$ using Eq. (\ref{eqn:conf esti}) \;
			Update $\Phi^* = \arg \min_{\Phi} \mathrm{L}_{\mathcal{I}} $ \;
		}
	}
	\caption{Building calibrated deep predictive models.}\label{algo}
\end{algorithm}

\noindent \textbf{Mean Estimation with Uncertainty Matching.}
\label{mean}
As formulated in Eq. (\ref{eqn:obj}), the goal is to leverage the calibrated intervals from the PI estimator, through uncertainty matching, to optimize for parameters $\Theta$. The $\mathcal{F}$ network is designed to predict the target along with uncertainty estimates, and the loss function $\mathrm{L}_{\mathcal{F}}$ is designed such that the fidelity of the predictions is measured with respect to the true targets, while the uncertainties are guided by the interval estimates from $\mathcal{I}$. We find that, by allowing the structure of estimated uncertainties to follow the pseudo-supervisory signal from $\mathcal{I}$, one can obtain higher quality estimates $\hat{\mathrm{y}}$. Now, we explain the two proposed strategies for uncertainty matching.

\noindent \textit{\textbf{a. Sigma Fit:}} In this variant, the predictive model $\mathcal{F}$ is designed as a heteroscedastic neural network, i.e. it returns mean and variance estimates $(\hat{\mathrm{y}}_i, \hat{\sigma}^2_i)$ for a given sample $\mathrm{x}_i$, and is optimized using the heteroscedastic loss~\cite{gal2016uncertainty}. Note that, this loss function is defined by assuming the predictions to follow a Gaussian distribution, and the estimated variances from the HNN can be used to estimate intervals for different levels of calibration $\alpha$,
\begin{equation}
\text{PI} = [\hat{\mathrm{y}} - z_{(1-\alpha)/2}\hat{\sigma}, \hat{\mathrm{y}} + z_{(1-\alpha)/2}\hat{\sigma}],
\end{equation} where $z$ indicates the $z-$score (i.e.\ $z_{(1-\alpha)/2} = 1.96,$ for $95\%$ calibration). We enforce the estimated $\hat{\sigma}$ to match the  appropriately scaled version of the interval width $w$, based on the calibration level achieved by $\mathcal{I}$ on the training dataset. Formally, the optimization objective can be written as:
\begin{equation} \label{bnn matching}
\mathrm{L}_{\mathcal{F}} =  \frac{1}{N} \sum_{i=1}^N \frac{\|\mathrm{y}_i - \mathcal{F}(\mathrm{x}_i;\Theta)\|_2^2}{2\sigma_i^2}  + \frac{1}{2}\log \sigma_i^2 + \lambda_m \|\sigma_i - \gamma \frac{w_i}{2}\|_1,
\end{equation}where $\lambda_m$ is a penalty term, the scale $\gamma = 1/z_{(1-\alpha_v)/2}$ with $\alpha_v$ being the calibration level achieved by $\mathcal{I}$.

\begin{figure}[t]
	\centering
	\includegraphics[width=1.\linewidth]{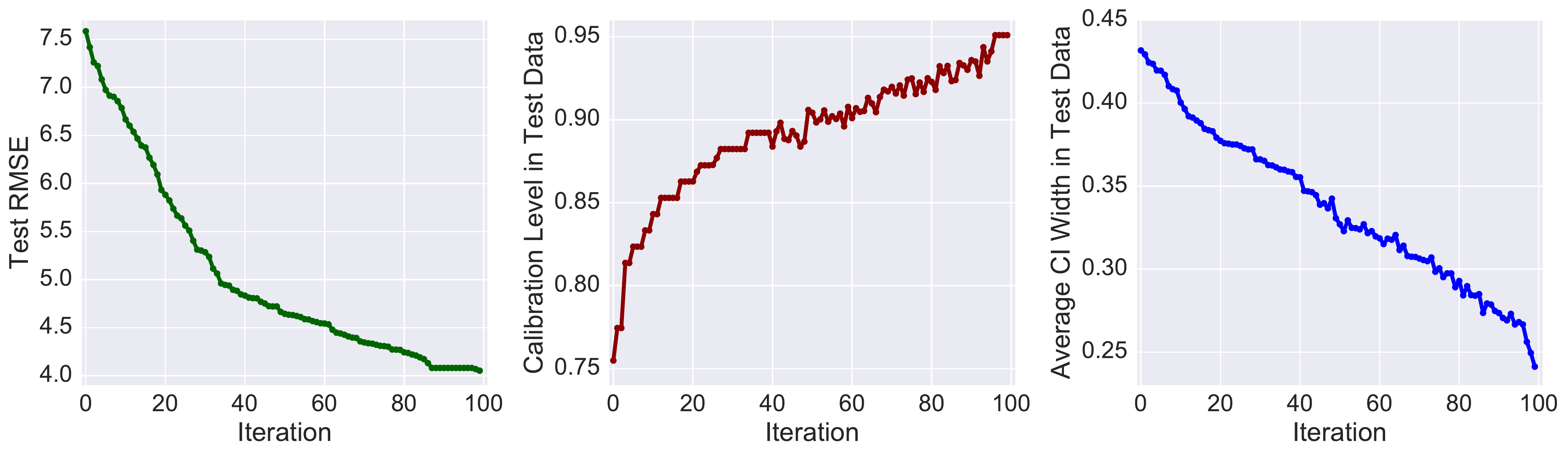}
	\includegraphics[width=1.\linewidth]{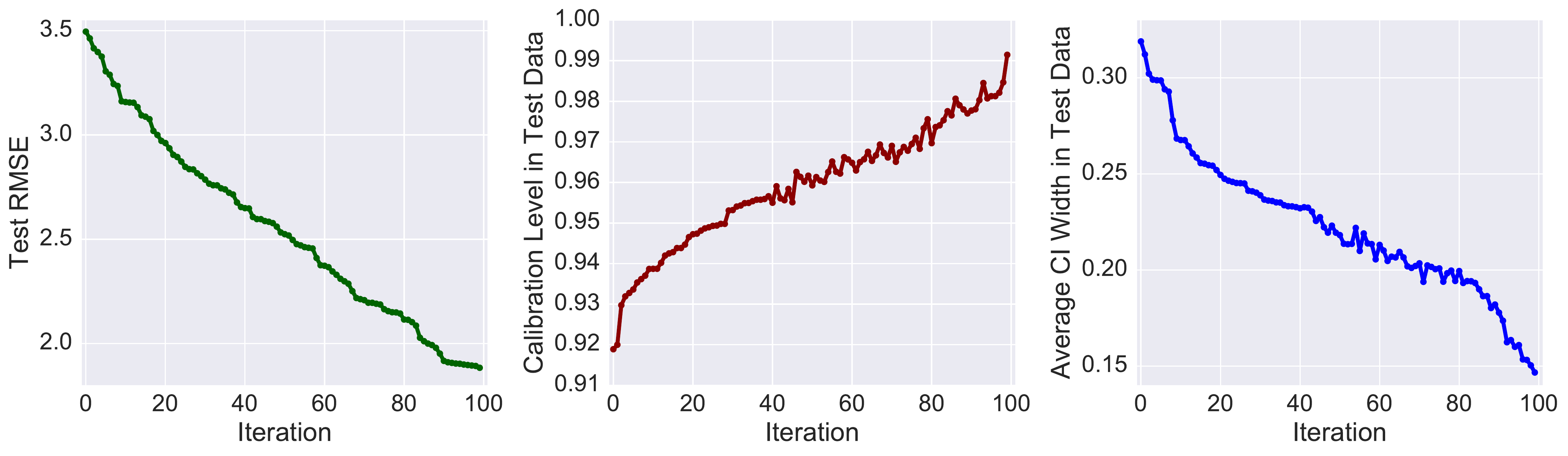}
	\caption{Convergence behavior of the proposed approach on the held-out test data for the Boston housing (top) and Parkinsons (bottom) datasets, for a target calibration level of $95\%$ and $99\%$ respectively : As the algorithm progresses, we observe that both the RMSE and calibration consistently improve. At the same time, the interval widths become sharp even while achieving high calibration.}
	\label{fig:boston-iter}
\end{figure}

\noindent \textit{\textbf{b. IQR Fit:}} In this strategy, we propose to obtain uncertainty estimates based on conditional quantiles of the target $\mathrm{y}$ and enforce them to match the intervals from the PI estimator. Let $F(\mathrm{y}) = P(\mathrm{y} \leq y)$ be the strictly monotonic cumulative distribution function of $\mathrm{y}$ assuming real values $y$ and $F^{-1}(\tau) = \inf \{y: F(\mathrm{y}=y) \geq \tau\}$ denotes the quantile distribution function for $0 \leq \tau \leq 1$. Quantile regression is aimed at inferring a desired quantile $\tau$ for the target $\mathrm{y}$, when the input $\mathrm{x} = x$, i.e., $F^{-1}(\tau | \mathrm{x} = x)$. This model can be estimated using the \textit{pinball} loss~\cite{koenker2001quantile}:
\begin{equation}
\mathrm{L}_{\tau}(\mathrm{y},\hat{\mathrm{y}}) =
\begin{cases}
\tau(\mathrm{y} - \hat{\mathrm{y}}) ,& \text{if }\mathrm{y} - \hat{\mathrm{y}} \geq 0, \\
(1-\tau)(\hat{\mathrm{y}} - \mathrm{y}),              & \text{otherwise}.
\end{cases}
\end{equation}Recently, \cite{tagasovska2018frequentist} showed that the \textit{inter-quantile range} (IQR) between upper and lower quantiles can produce calibrated estimates of the aleatoric uncertainties. Hence, we propose an uncertainty estimator similar to~\cite{tagasovska2018frequentist}, where the $\alpha$ prediction interval can be obtained around the median as:
\begin{align}
&\text{IQR}(\mathrm{x}; \tau^u, \tau^l) = \mathcal{F}(\mathrm{x}|\tau = \tau^u, \Theta) - \mathcal{F}(\mathrm{x}|\tau = \tau^l, \Theta), \\
\nonumber&\tau^u = (1 + \alpha)/2, \tau^l = (1 - \alpha)/2.
\end{align}Given the interval estimates from the PI network, the loss function $\mathrm{L}_{\mathcal{F}}$ can be constructed as:
\begin{align} \label{IQR matching}
\nonumber \mathrm{L}_{\mathcal{F}} &=  \frac{1}{N} \sum_{i=1}^N \|\mathrm{y}_i - \mathcal{F}(\mathrm{x}_i; \Theta)\|_2^2 + \lambda_u \mathrm{L}_{\tau^u}(\mathrm{y}_i, \mathcal{F}(\mathrm{x}_i; \Theta)) \\
& + \lambda_l \mathrm{L}_{\tau^l}(\mathrm{y}_i, \mathcal{F}(\mathrm{x}_i; \Theta)) + \lambda_m \|\text{IQR}(\mathrm{x}_i) - w_i \|_1.
\end{align}

\subsection{Algorithm}
The overall algorithm, which repeats the regularized mean estimation and the interval prediction in an alternating fashion, is provided in Algorithm~\ref{algo}. The mean estimator takes the intervals into account for its update. Similarly, we include the residual structure from the mean estimator for interval estimation. For a fixed interval, the improved mean estimate can increase the calibration error (CE) by achieving a higher likelihood even for smaller $\alpha$. However, when reupdating the interval estimator, we expect the widths to become sharper to reduce CE. This collaborative process thus leads to superior quality mean estimates and highly calibrated intervals. While each of the steps are run for $n_m$ and $n_c$ epochs respectively, the outer loop is executed until convergence. In all our experiments, both hyper-parameters $n_m$ and $n_c$ are set at $10$ and the algorithm returns optimized parameter estimates for $\mathcal{F}$ and $\mathcal{I}$. It is important to note that, we do not require any additional recalibration step for the validation/testing stages, in contrast to approaches such as~\cite{kuleshov2018accurate,platt1999probabilistic}. Figure~\ref{fig:boston-iter} illustrates the convergence behavior of the proposed approach for a regression task on two datasets. We observe that, as the algorithm progresses, the proposed approach produces well-calibrated intervals that are also sharp (measured using the average of interval widths $w_i$). More importantly, the uncertainty matching based regularization produces much improved quality predictors, as evidenced by the improvements to the prediction quality (measured using RMSE).

\begin{table*}[t]
	\centering
	\caption{Test RMSE from the regression experiments on $6$ UCI benchmark datasets. For all cases, the results showed are for the case of $\alpha = 0.9$, i.e. $90\%$ calibration. The results reported were obtained by averaging over $5$ random trials.}

	\begin{tabular}{|c|c|c|c|c|c|c|}
		\hline
		\rule{0pt}{2ex} 
		\textbf{Method} & \cellcolor{gray!5}\textbf{Crime} & \cellcolor{gray!10}\textbf{Red Wine} & \cellcolor{gray!15}\textbf{White Wine} & \cellcolor{gray!20}\textbf{Parkinsons} & \cellcolor{gray!25}\textbf{Boston} & \cellcolor{gray!30}\textbf{Auto MPG} \\ \hline \hline
        \rule{0pt}{2ex}   
		MC Dropout & \cellcolor{gray!5}\textbf{0.139} & \cellcolor{gray!10}0.686 & \cellcolor{gray!15}0.748 &  \cellcolor{gray!20}2.78 & \cellcolor{gray!25}5.51 & \cellcolor{gray!30}4.61 \\
        \rule{0pt}{2ex}   
		Concrete Dropout & \cellcolor{gray!5}0.144 & \cellcolor{gray!10}0.745 & \cellcolor{gray!15}0.764 &  \cellcolor{gray!20}3.1 &\cellcolor{gray!25} 5.88 & \cellcolor{gray!30}5.19 \\
        \rule{0pt}{2ex} 
		Quantile Estimator & \cellcolor{gray!5}0.152 & \cellcolor{gray!10}0.737 & \cellcolor{gray!15}0.782 &  \cellcolor{gray!20}2.91 &\cellcolor{gray!25} 4.802 & \cellcolor{gray!30}4.32\\
        \rule{0pt}{2ex} 
		BNN & \cellcolor{gray!5}0.159 & \cellcolor{gray!10}0.823 & \cellcolor{gray!15}0.871 &  \cellcolor{gray!20}2.493 & \cellcolor{gray!25}4.025 & \cellcolor{gray!30}\textbf{3.07} \\
        \rule{0pt}{2ex} 
		HNN & \cellcolor{gray!5}\textbf{0.138} & \cellcolor{gray!10}0.714 & \cellcolor{gray!15}0.757 &  \cellcolor{gray!20}2.842 & \cellcolor{gray!25}4.762 & \cellcolor{gray!30}3.98 \rule{0pt}{2ex}  \\
		\hline
		\hline
		Sigma Fit & \cellcolor{gray!5}\textbf{0.138} & \cellcolor{gray!10}0.599 & \cellcolor{gray!15}0.702 &  \cellcolor{gray!20}\textbf{2.017} & \cellcolor{gray!25}4.088 & \cellcolor{gray!30}3.23 \\
        \rule{0pt}{2ex} 
		IQR Fit & \cellcolor{gray!5}\textbf{0.138} & \cellcolor{gray!10}\textbf{0.581} & \cellcolor{gray!15}\textbf{0.691} &  \cellcolor{gray!20}2.206 &\cellcolor{gray!25} \textbf{4.011} & \cellcolor{gray!30}3.18\\

		\hline
	\end{tabular}
	\label{table:regression exp}
\end{table*}

\begin{figure*}[t]
	\centering
	\begin{subfigure}[b]{0.48\textwidth}
	        \centering
			\includegraphics[width=0.48\linewidth, keepaspectratio=true]{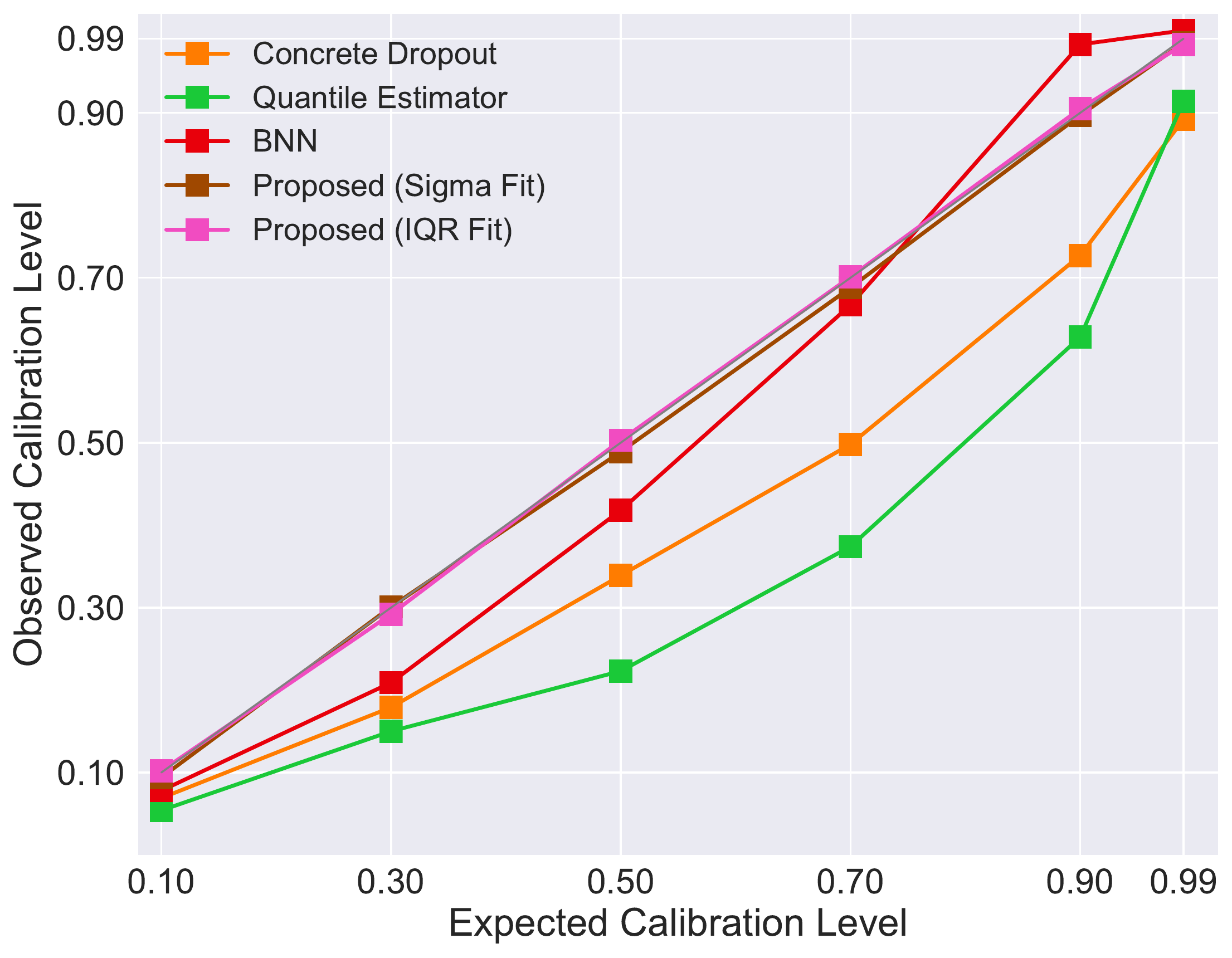}
			\includegraphics[width=0.48\linewidth, keepaspectratio=true]{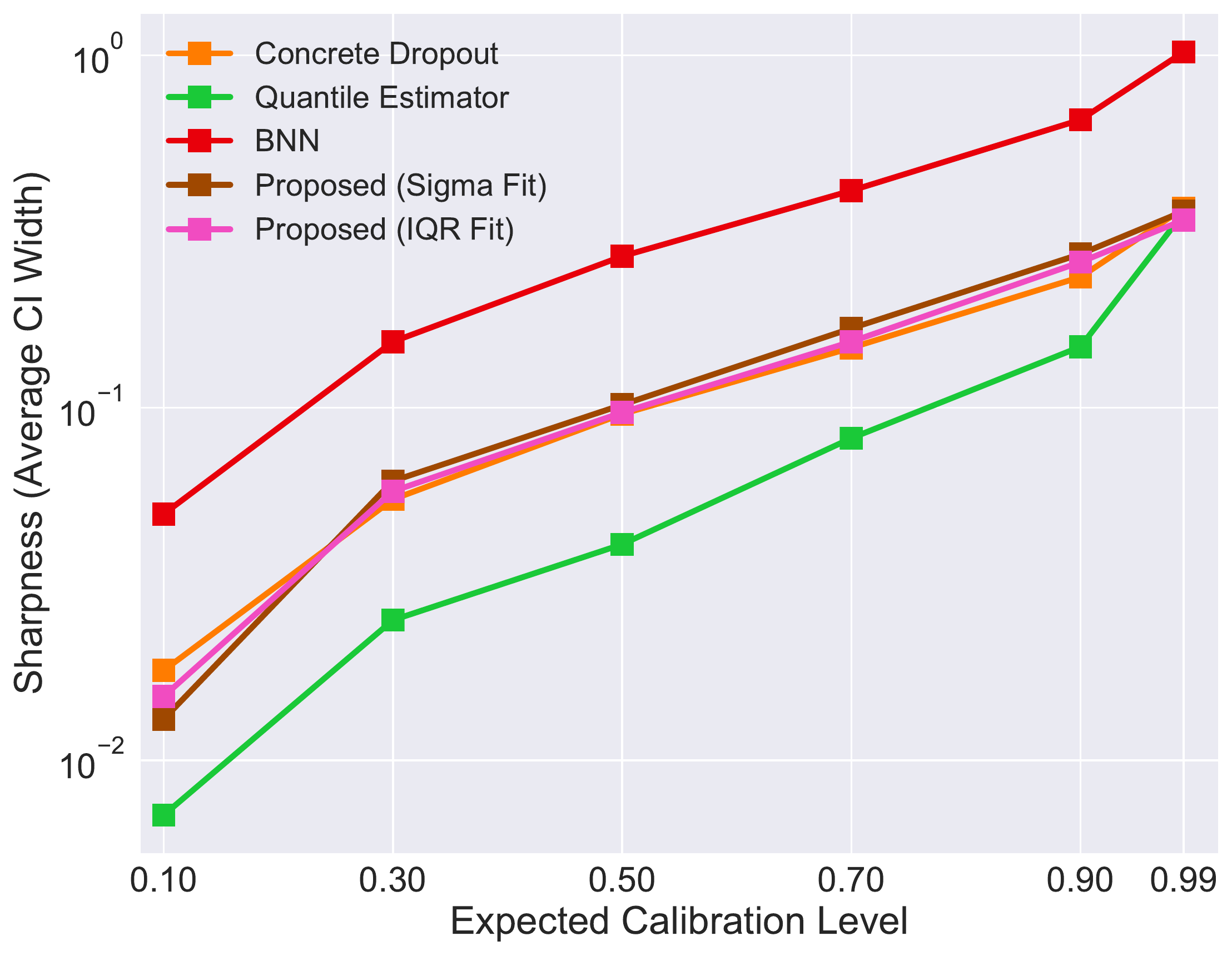}
		\caption{Boston Housing}
	    \label{fig:boston}		
    \end{subfigure}
    \begin{subfigure}[b]{0.48\textwidth}
	        \centering
			\includegraphics[width=0.48\linewidth, keepaspectratio=true]{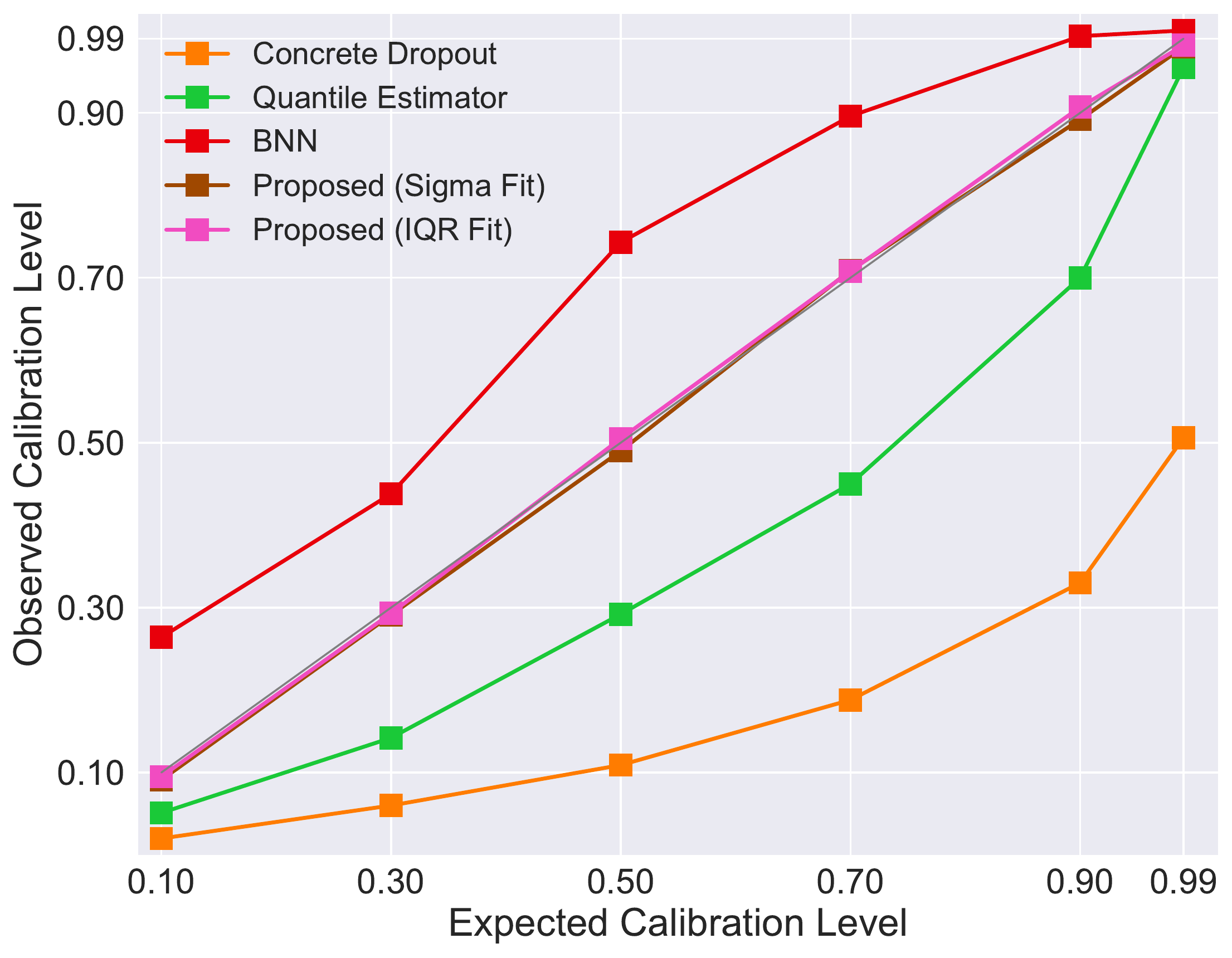}
			\includegraphics[width=0.48\linewidth, keepaspectratio=true]{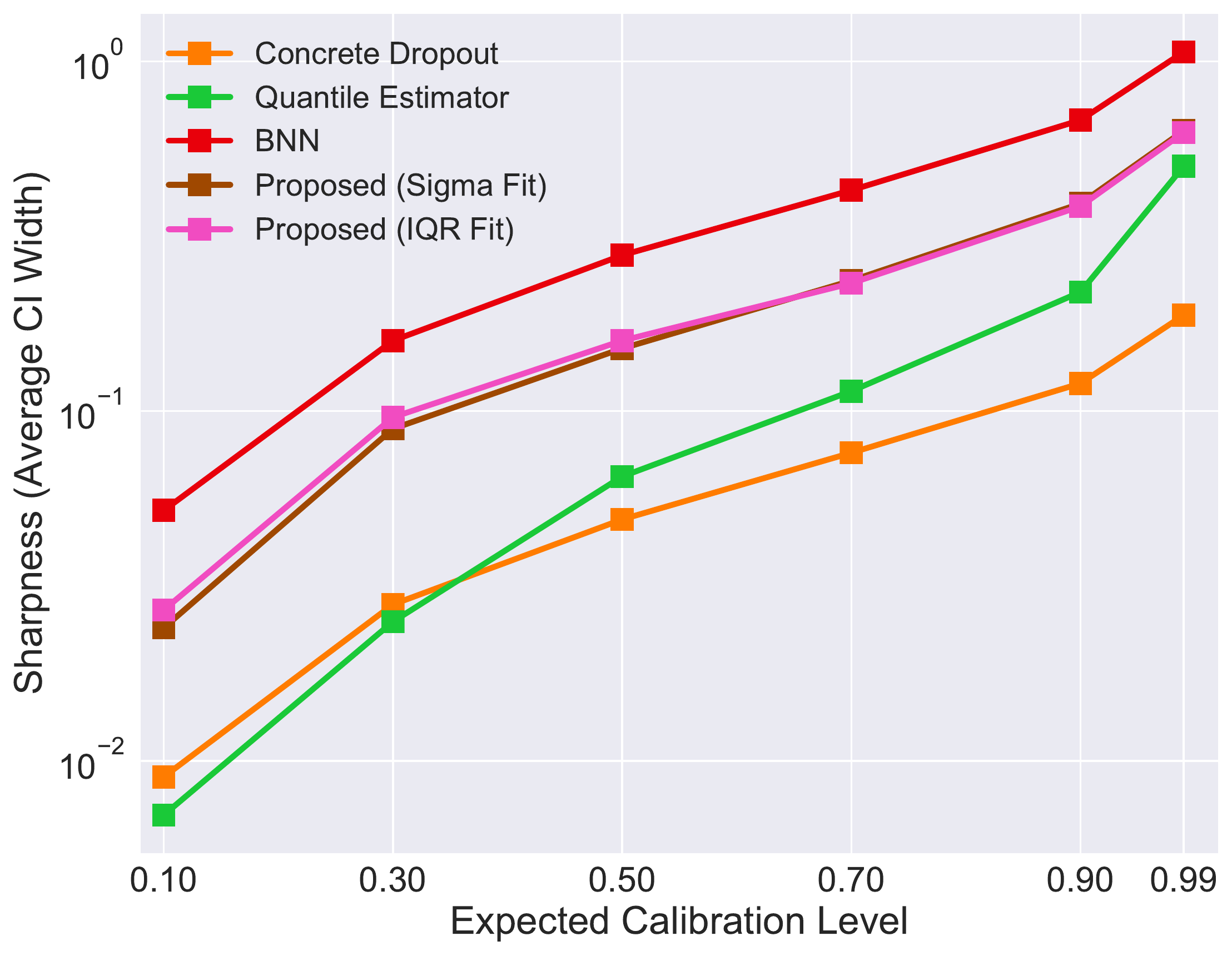}
		\caption{White Wine}
	    \label{fig:whitewine}		
    \end{subfigure}
    
    \vfill
    
    \begin{subfigure}[b]{0.48\textwidth}
	        \centering
			\includegraphics[width=0.48\linewidth, keepaspectratio=true]{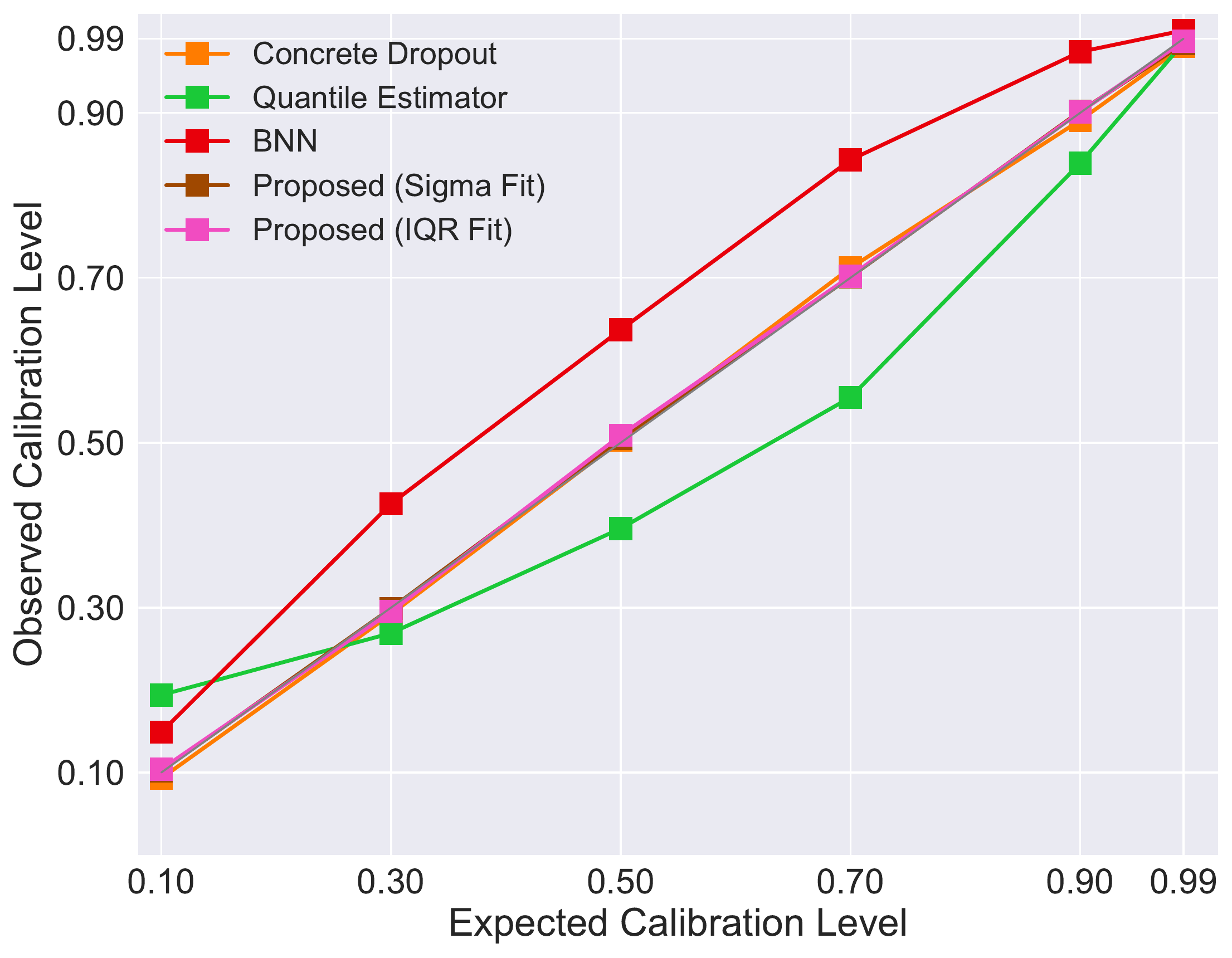}
			\includegraphics[width=0.48\linewidth, keepaspectratio=true]{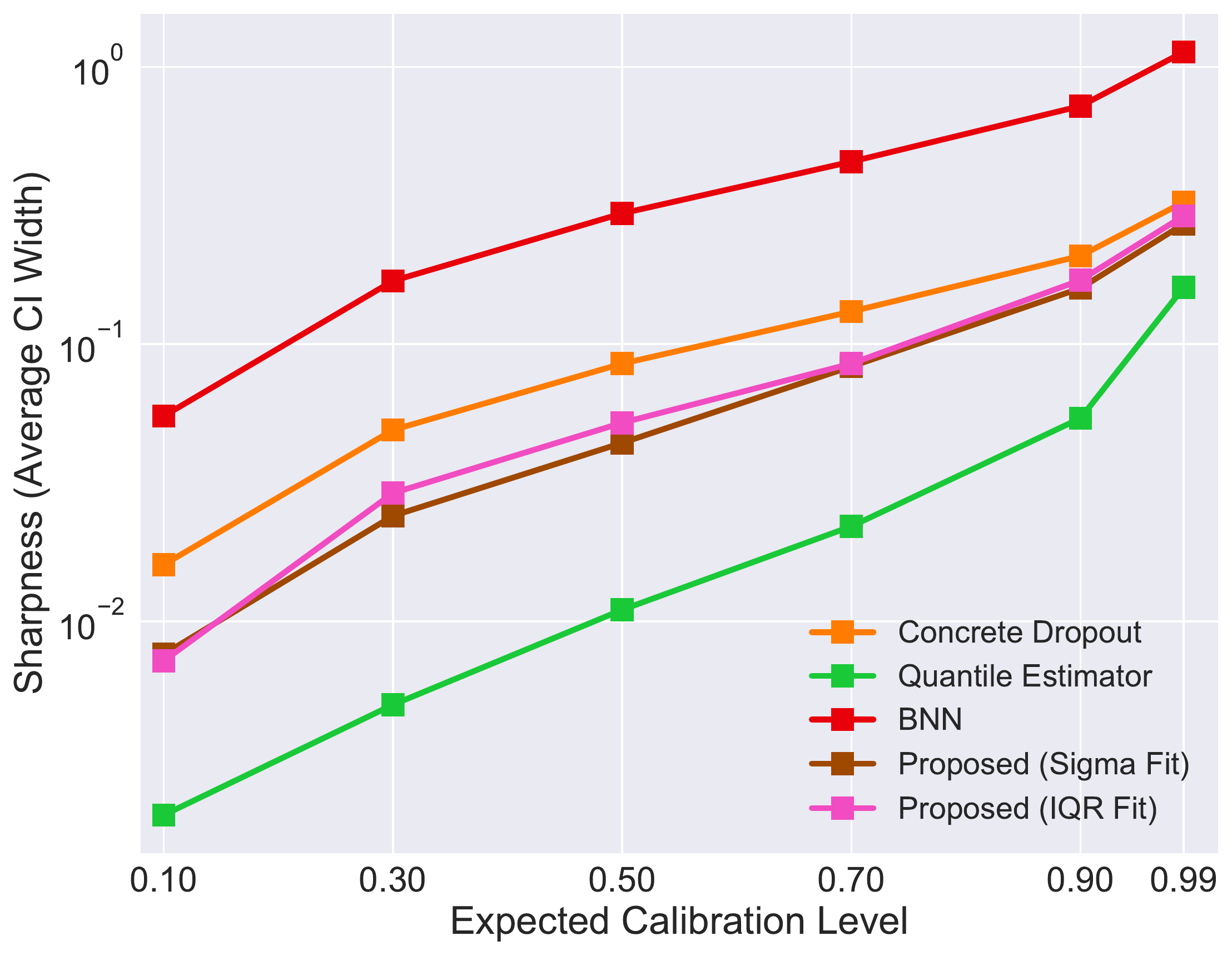}
		\caption{Parkinsons}
	    \label{fig:parkinsons}		
    \end{subfigure}
    \begin{subfigure}[b]{0.48\textwidth}
	        \centering
			\includegraphics[width=0.48\linewidth, keepaspectratio=true]{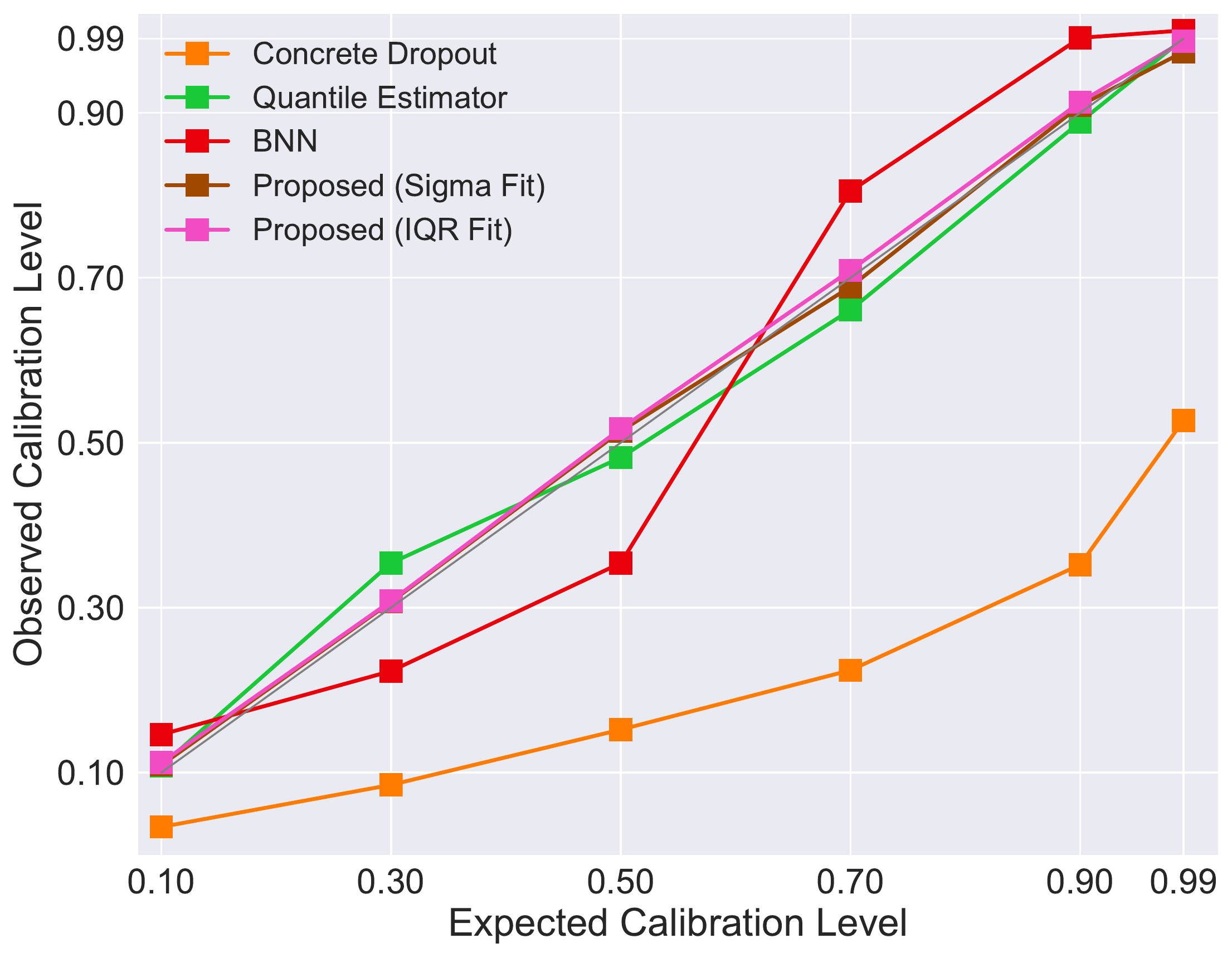}
			\includegraphics[width=0.48\linewidth, keepaspectratio=true]{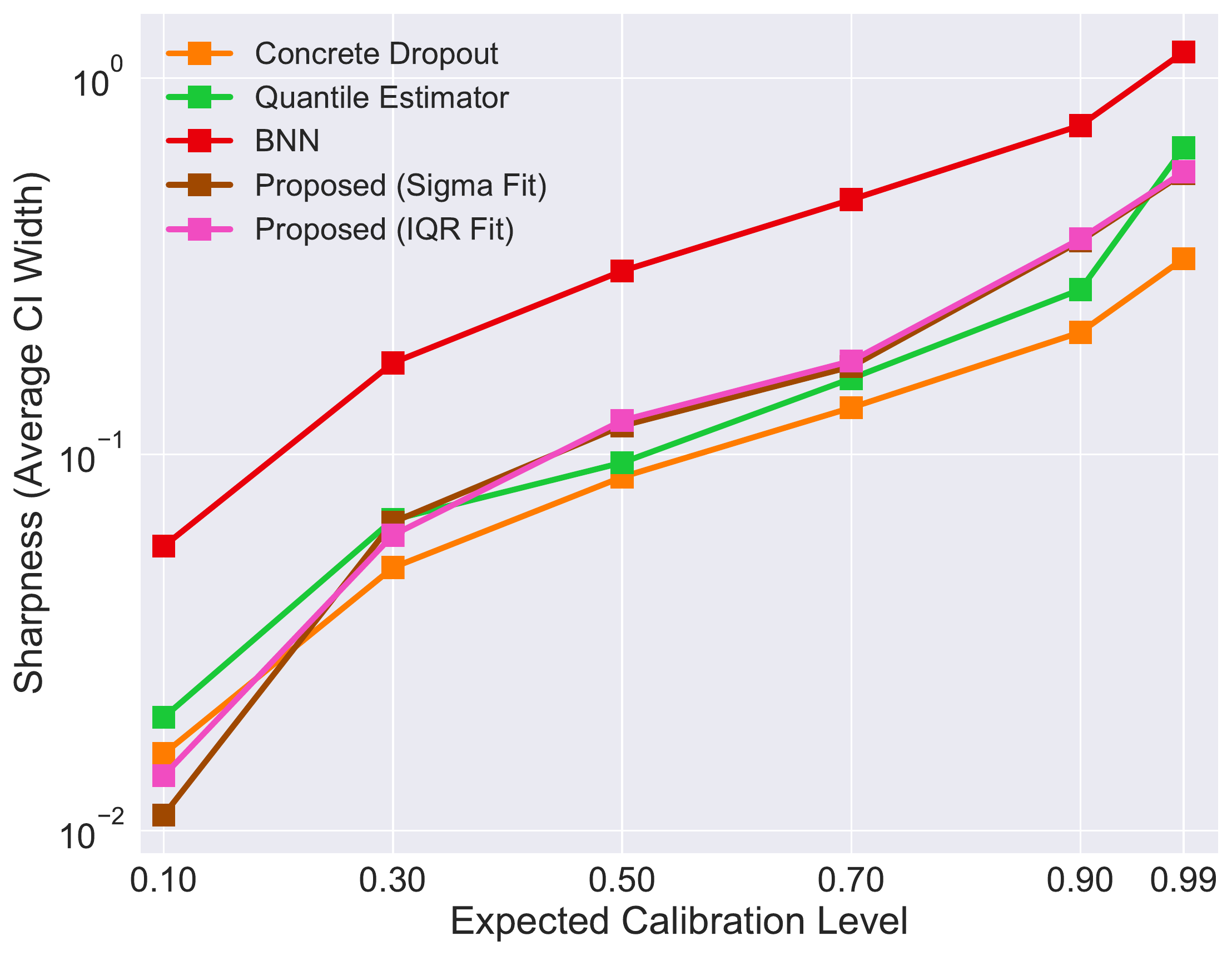}
		\caption{Auto MPG}
	    \label{fig:autompg}		
    \end{subfigure}
	\caption{Evaluating the quality of estimated intervals - For each dataset, we plot the expected calibration vs observed calibration on test data to demonstrate the generality of the proposed PI estimator, when compared to baselines. Further, we plot expected calibration vs average PI width to show how the estimated intervals compare to existing uncertainty estimators.}
	\label{fig:regression}
\end{figure*}

\section{Empirical Studies}
In this section, we present empirical studies to compare the proposed approach against popularly adopted methods for producing prediction intervals. More specifically, we considered three different use cases where the goal was to predict continuous valued target variables --  regression, forecasting with time-series data and object localization in images (i.e. predicting bounding boxes).

\begin{table*}[t]
	\centering
	\caption{Results from the time-series forecasting experiments - We report the test RMSE, calibration error and average width metrics for each of the cases, when $\alpha$ was set to $0.95$.}
	\begin{tabular}{|c|c|c|c|c|c|c|c|}
		\hline
		\rule{0pt}{2ex} 
		\textbf{Method} & \textbf{Metric} & \cellcolor{gray!5}\textbf{NSW2013} & \cellcolor{gray!10}\textbf{Bike Sharing} & \cellcolor{gray!15}\textbf{TAS2016} & \cellcolor{gray!20} \textbf{Traffic} & \cellcolor{gray!25}\textbf{Beijing} & \cellcolor{gray!30}\textbf{Air Quality} \\ \hline \hline
        \rule{0pt}{2ex} 
		\multirow{3}{*}{MC Dropout } & RMSE & \cellcolor{gray!5}125.48 & \cellcolor{gray!10} 62.68 & \cellcolor{gray!15}22.12 & \cellcolor{gray!20}2.03E+08 &\cellcolor{gray!25} 22.42 & \cellcolor{gray!30}0.66 \\
        \rule{0pt}{2ex} 
		& $\text{CE}_{0.95}$ & \cellcolor{gray!5}0.04 & \cellcolor{gray!10}0.30 & \cellcolor{gray!15}0.196 & \cellcolor{gray!20}0.037 & \cellcolor{gray!25}0.177 & \cellcolor{gray!30}0.02 \\
        \rule{0pt}{2ex} 
		& $\text{AW}_{0.95}$ & \cellcolor{gray!5}587.54 & \cellcolor{gray!10}90.94 & \cellcolor{gray!15}50.37 & \cellcolor{gray!20}8.39E+08 & \cellcolor{gray!25}38.53 &\cellcolor{gray!30} 2.52 \\ \hline \hline
        \rule{0pt}{2ex} 
		\multirow{3}{*}{ Concrete Dropout } & RMSE & \cellcolor{gray!5}153.16 & \cellcolor{gray!10}64.81 & \cellcolor{gray!15}23.11 & \cellcolor{gray!20}2.25E+08 & \cellcolor{gray!25}22.46 & \cellcolor{gray!30}0.65 \\
        \rule{0pt}{2ex} 
		& $\text{CE}_{0.95}$ & \cellcolor{gray!5}0.046 & \cellcolor{gray!10}0.277 & \cellcolor{gray!15}0.145 &\cellcolor{gray!20} 0.073 & \cellcolor{gray!25}0.12 & \cellcolor{gray!30}0.105 \\
        \rule{0pt}{2ex} 
		& $\text{AW}_{0.95}$ & \cellcolor{gray!5}696.19 & \cellcolor{gray!10}96.67 & \cellcolor{gray!15}53.58 & \cellcolor{gray!20}8.93E+08 & \cellcolor{gray!25}48.42 & \cellcolor{gray!30}2.73 \\ \hline \hline
        \rule{0pt}{2ex} 
		\multirow{3}{*}{Quantile Estimator} & RMSE & \cellcolor{gray!5}131.18 & \cellcolor{gray!10}65.19 & \cellcolor{gray!15}23.44 & \cellcolor{gray!20}2.07E+08 & \cellcolor{gray!25}22.45 & \cellcolor{gray!30}0.77 \\
        \rule{0pt}{2ex} 
		& $\text{CE}_{0.95}$ & \cellcolor{gray!5}0.07 & \cellcolor{gray!10}0.22 & \cellcolor{gray!15}0.19 &\cellcolor{gray!20} 0.07 & \cellcolor{gray!25}0.11 & \cellcolor{gray!30}0.241 \\
        \rule{0pt}{2ex} 
		& $\text{AW}_{0.95}$ & \cellcolor{gray!5}533.31 & \cellcolor{gray!10}99.34 & \cellcolor{gray!15}49.17 & \cellcolor{gray!20}8.6E+08 & \cellcolor{gray!25}49.17 & \cellcolor{gray!30}2.71 \\ \hline \hline
        \rule{0pt}{2ex} 
		\multirow{3}{*}{BNN} & RMSE & \cellcolor{gray!5}153.16 & \cellcolor{gray!10}64.81 & \cellcolor{gray!15}23.11 & \cellcolor{gray!20}2.25E+08 & \cellcolor{gray!25}22.46 & \cellcolor{gray!30}0.65 \\
        \rule{0pt}{2ex} 
		& $\text{CE}_{0.95}$ & \cellcolor{gray!5}0.046 & \cellcolor{gray!10}0.277 & \cellcolor{gray!15}0.145 &\cellcolor{gray!20} 0.073 & \cellcolor{gray!25}0.12 & \cellcolor{gray!30}0.105 \\
        \rule{0pt}{2ex} 
		& $\text{AW}_{0.95}$ & \cellcolor{gray!5}696.19 & \cellcolor{gray!10}96.67 & \cellcolor{gray!15}53.58 & \cellcolor{gray!20}8.93E+08 & \cellcolor{gray!25}48.42 & \cellcolor{gray!30}2.73 \\ \hline \hline
        \rule{0pt}{2ex} 
		\multirow{3}{*}{ HNN } & RMSE & \cellcolor{gray!5}124.59 & \cellcolor{gray!10}64.38 & \cellcolor{gray!15}22.81 & \cellcolor{gray!20}1.36E+08 & \cellcolor{gray!25}22.83 & \cellcolor{gray!30}0.72 \\
        \rule{0pt}{2ex} 
		& $\text{CE}_{0.95}$ & \cellcolor{gray!5}0.008 & \cellcolor{gray!10}0.004 & \cellcolor{gray!15}0.013 & \cellcolor{gray!20}0.014 &\cellcolor{gray!25} 0.028 & \cellcolor{gray!30}0.041 \\
        \rule{0pt}{2ex} 
		& $\text{AW}_{0.95}$ & \cellcolor{gray!5}452.05 & \cellcolor{gray!10}199.64 & \cellcolor{gray!15}91.24 & \cellcolor{gray!20}5.62E+08 & \cellcolor{gray!25}98.95 & \cellcolor{gray!30}6.16 \\\hline \hline
        
        \rule{0pt}{2ex} 
		\multirow{3}{*}{Sigma Fit } & RMSE & \cellcolor{gray!5}\textbf{90.31} & \cellcolor{gray!10}58.97 &\cellcolor{gray!15} 17.95 &\cellcolor{gray!20} 1.27E+08 & \cellcolor{gray!25}\textbf{18.65} & \cellcolor{gray!30}\textbf{0.53} \\
        \rule{0pt}{2ex} 
		& $\text{CE}_{0.95}$ &\cellcolor{gray!5} \textbf{0.002} & \cellcolor{gray!10}0.005 & \cellcolor{gray!15}0.009 & \cellcolor{gray!20}0.006 & \cellcolor{gray!25}0.011 & \cellcolor{gray!30}0.007 \\
        \rule{0pt}{2ex} 
		& $\text{AW}_{0.95}$ & \cellcolor{gray!5}424.59 & \cellcolor{gray!10}169.81 & \cellcolor{gray!15}77.71 & \cellcolor{gray!20}5.49E+08 & \cellcolor{gray!25}88.19 & \cellcolor{gray!30}2.99 \\\hline \hline
        \rule{0pt}{2ex} 
		\multirow{3}{*}{IQR Fit } & RMSE & \cellcolor{gray!5}94.85 & \cellcolor{gray!10}\textbf{53.65} &\cellcolor{gray!15} \textbf{17.64} &\cellcolor{gray!20} \textbf{1.21E+08} & \cellcolor{gray!25}19.43 & \cellcolor{gray!30}0.62 \\
        \rule{0pt}{2ex} 
		& $\text{CE}_{0.95}$ & \cellcolor{gray!5}{0.003} & \cellcolor{gray!10}\textbf{0.002} & \cellcolor{gray!15}\textbf{0.005} & \cellcolor{gray!20}\textbf{0.003} & \cellcolor{gray!25}\textbf{0.01} & \cellcolor{gray!30}\textbf{0.004} \\
        \rule{0pt}{2ex} 
		& $\text{AW}_{0.95}$ &\cellcolor{gray!5}{421.63} & \cellcolor{gray!10}182.34 & \cellcolor{gray!15}80.44 &\cellcolor{gray!20} 5.43E+08 & \cellcolor{gray!25}87.48 & \cellcolor{gray!30}3.11 \\
		\hline
	\end{tabular}
	\label{tab:ts}
\end{table*}

\begin{figure*}[t]
	\centering
	\begin{subfigure}[b]{0.24\textwidth}
	        \centering
			\includegraphics[width=\linewidth, keepaspectratio=true]{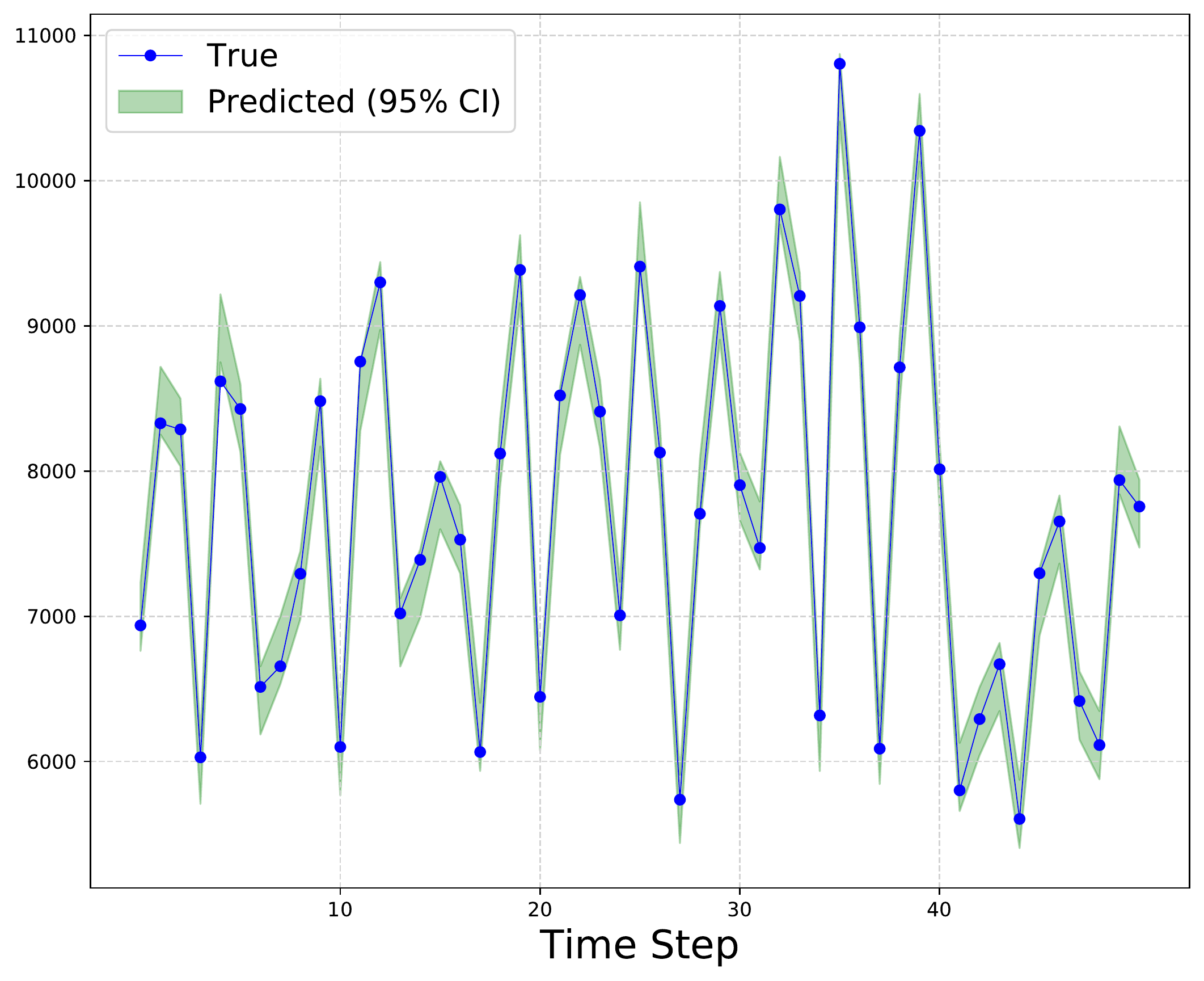}
		\caption{NSW2016}
	    \label{fig:nsw}		
    \end{subfigure}
    \begin{subfigure}[b]{0.235\textwidth}
	        \centering
			\includegraphics[width=\linewidth, keepaspectratio=true]{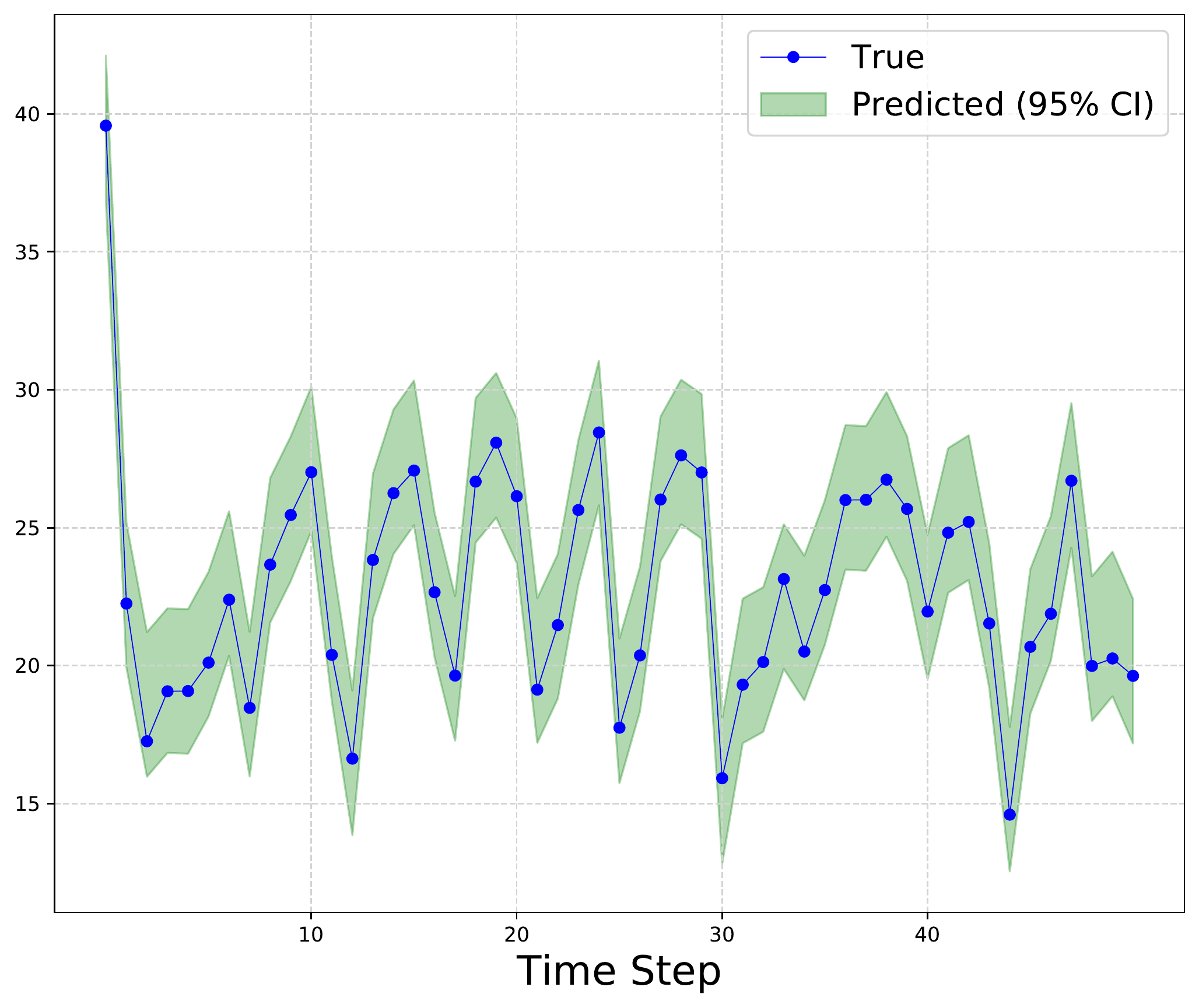}
		\caption{Air Quality}
	    \label{fig:pollution}		
    \end{subfigure}
    \begin{subfigure}[b]{0.24\textwidth}
	        \centering
			\includegraphics[width=\linewidth, keepaspectratio=true]{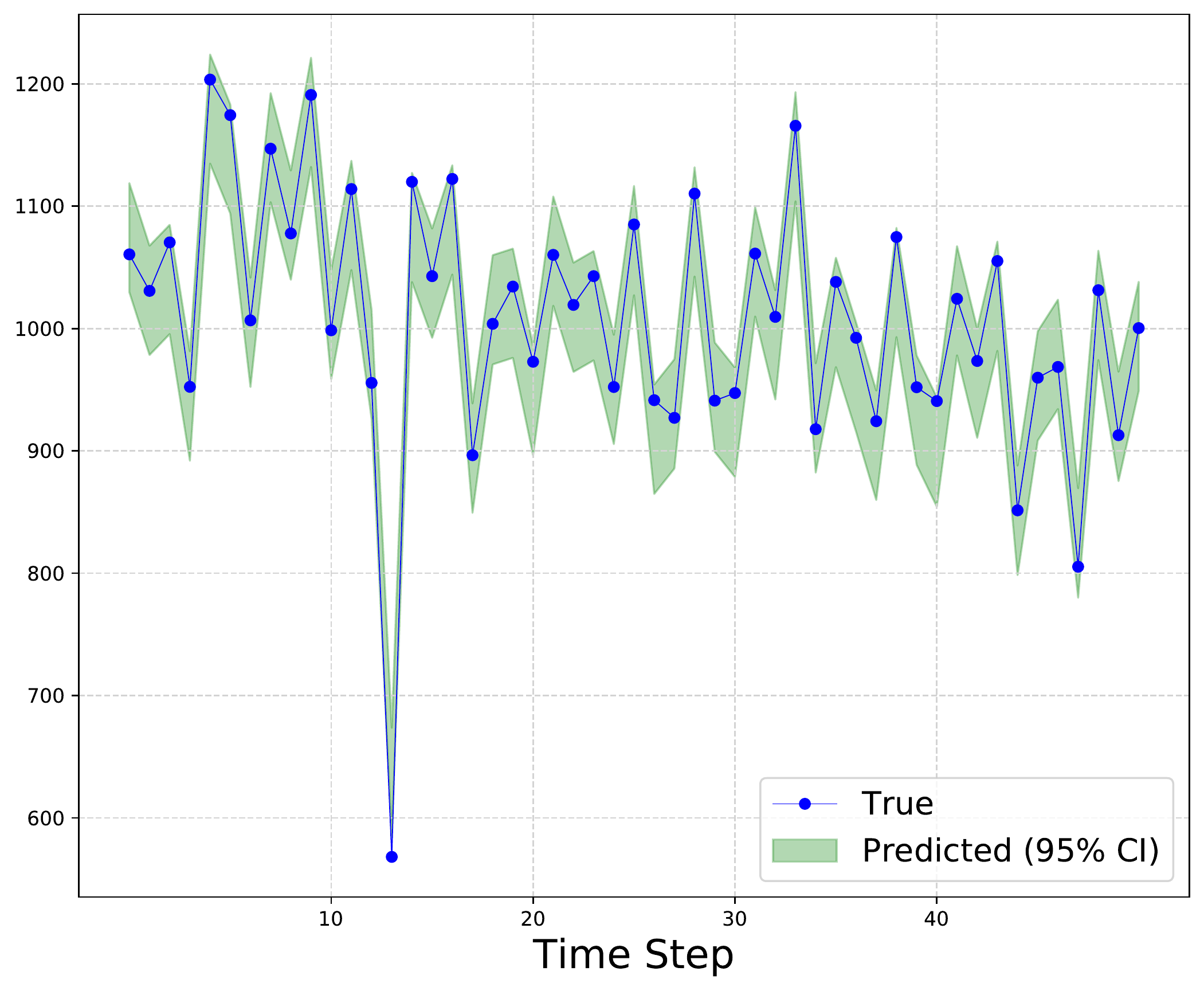}
		\caption{TAS2016}
	    \label{fig:tas2016}		
    \end{subfigure}
    \begin{subfigure}[b]{0.24\textwidth}
	        \centering
			\includegraphics[width=\linewidth, keepaspectratio=true]{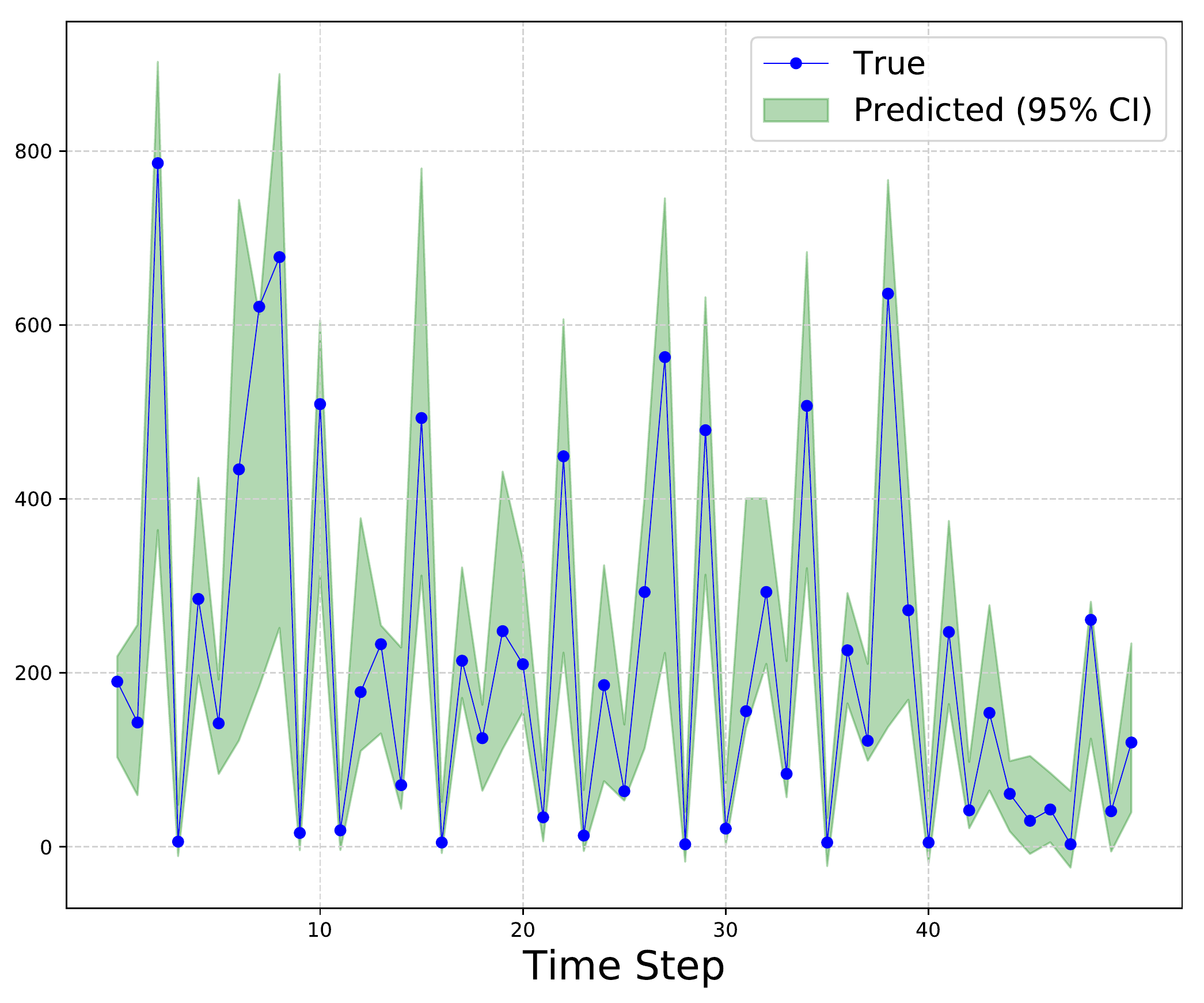}
		\caption{Bike Sharing}
	    \label{fig:bike}		
    \end{subfigure}
\caption{\textit{Forecasting}: Prediction intervals obtained using the proposed \textit{IQR Fit} method for the calibration level $\alpha = 0.95$ on the test data. It can be seen that in all cases, our method produces sharp intervals that are well calibrated.}
\label{fig:ts}
\end{figure*}

\subsection{Experiment Setup}
\noindent \textbf{Hyper-parameter Choices}: We used Algorithm 1 to solve the optimization problem in Eq. (\ref{eqn:obj}), based on the two proposed strategies, namely \textit{Sigma Fit} and \textit{IQR Fit}. We used the following hyper-parameter choices for all experiments: The penalties for the loss $\mathcal{L}_{\mathcal{I}}$ in Eq. (\ref{eqn:conf esti}) were fixed at $\beta_n = 0.1$ and $\beta_s = 0.3$ respectively. For the \textit{Sigma Fit} method, the penalty for the uncertainty matching term was set at $\lambda_m = 0.5$. Similarly, the hyper-parameters for constructing the loss $\mathcal{L}_{\mathcal{F}}$ for \textit{IQR Fit}, we used $\lambda_m = 0.4, \lambda_u = \lambda_l = 0.3$. While the outer loop in Algorithm 1 was run until convergence, both the mean estimator and the PI estimator networks were trained for $n_m = n_c = 10$ epochs in each iteration.

\noindent \textbf{Evaluation Metrics}: For both regression and time-series forecasting applications, we used the standard root mean square error (RMSE) metric to evaluate the quality of mean predictions. In the case of object localization, we used the intersection-over-union (IoU) metric, that measures the ratio between the area under intersection of the true and predicted bounding boxes, and the area under union of both bounding boxes. Since our primary objective is to produce calibrated estimators, we evaluate the quality of prediction intervals on the held-out validation set using the calibration error metric, $CE_{\alpha}$, which is the same as the $\mathrm{L}_{emce}$ term in Eq. (\ref{eqn:emce}). In order to ensure that the PI estimator network does not produce a degenerate solution (arbitrarily large intervals that are always calibrated), we included the $\mathrm{L}_{sharp}$ term in Eq. (\ref{eqn:conf esti}). Hence, we also evaluate how well that constraint is met using the average width metric: $AW_{\alpha} =  \frac{1}{N}{ \sum_{i=1}^{N}(\delta_i^l+\delta_i^u)}$.

\noindent \textbf{Baselines}: We compare against the following baseline techniques that are commonly adopted to produce prediction intervals, in lieu of simple point-estimates. Though there exists an entire class of recalibration methods~\cite{kuleshov2018accurate}, which utilizes an additional recalibration dataset to refine the learned estimator with a calibration objective, our focus is on single-shot calibration methods.

\noindent \textit{MC Dropout}~\cite{gal2016dropout}:
In this approach, dropout in deep neural networks is cast as an approximate Bayesian inference in deep Gaussian processes. Using dropout during both train and test phases simulates Monte-Carlo sampling from the network with non-deterministic weights, thus capturing epistemic uncertainties. For all experiments, the dropout probability was set at $0.5$.

\noindent \textit{Concrete Dropout}~\cite{gal2017concrete}:
This was designed to automatically tune the dropout probability based on data characteristics, and hence it is expected to produce better calibrated intervals in comparison to MC Dropout.

\noindent \textit{Bayesian Neural Networks (BNN)}~\cite{blundell2015weight}:
In this approach, the underlying distribution on network weights are approximated with a Gaussian (variational approximation), and the posterior inference based on observed data is then carried out by sampling from this approximate distribution, thus capturing epistemic uncertainties.

\noindent \textit{Quantile Estimator}~\cite{tagasovska2018frequentist}:
Inter-quantile range from conditional quantiles of predictors can be used to estimate the aleatoric uncertainties and hence prediction intervals for a given $\alpha$ confidence can be obtained by choosing appropriate upper and lower quantiles.

\noindent \textit{Heteroscedastic Neural Networks (HNN)}~\cite{cvuncertainties}:
Using the Gaussian likelihood formulation, the  data dependent observation noise, i.e. aleatoric uncertainties, can be captured through the heteroscedastic loss.

\subsection{Regression using Fully Connected Networks}
In this experiment, we considered the problem of regressing to a continuous target variable from high-dimensional features. For each dataset, we used a random $80/20$ split for train/test, and we report the averaged results obtained from $5$ random trials. As a preprocessing step, we rescaled the target data into the range $[0,1]$ before training the models.

\noindent \textbf{Datasets}: We considered $6$ datasets from the UCI repository~\cite{Dua:2019}, which ranged between $7$ and $124$ in their dimensionality, and between $391$ and $4898$ in terms of sample size: crime, red wine quality, white wine quality, parkinsons, boston housing and auto mpg.

\noindent \textbf{Model Architecture}: For all approaches, we used a fully connected network with $5$ linear layers, each followed by a ReLU activation function, and the MSE loss for training.

\noindent \textbf{Results}: A meaningful interval is one that reflects the informative uncertainties in data or modeling process and is well calibrated. Hence, we measure the quality of calibration achieved using the proposed approach at different calibration levels $\alpha$ - (i) \textit{expected calibration} vs \textit{observed calibration}; and (ii) \textit{expected calibration} vs \textit{average PI width}. While the former evidences the generality of the PI estimator to unseen test data, the latter shows how the estimated intervals compare to conventional uncertainty estimators. Similar to the findings of existing studies, we also find that existing approaches such as MC Dropout, Concrete Dropout and BNN are not well calibrated. In comparison, we found the heteroscedastic neural networks to be better calibrated. Figure \ref{fig:regression} shows the results for $4$ different datasets (rest in supplementary) and we find that the proposed approach achieves significantly improved calibration, when compared to the baseline methods. Interestingly, from the \textit{average PI width} plots, we find that the estimated intervals are typically larger than the \textit{aleatoric} uncertainties from the quantile estimator, while being much lower than the estimates from BNNs. Finally, from Table \ref{table:regression exp}, it is apparent the \textit{uncertainty matching} process leads to improved regularization, thus producing higher-quality mean estimates.

\subsection{Time-series Forecasting using LSTMs}
In this study, we consider the problem of forecasting in time-varying data by leveraging information from the past samples. More specifically, we attempt to look-ahead by one sample $\mathrm{x}[t+1]$ using the observations $\mathrm{x}[t-k:t+1]$.

\noindent \textbf{Datasets}: We considered $6$ time-series datasets: NSW2013 and TAS2016 annual electricity demand recordings from the Australian Energy Market Operator; bike sharing, air quality, Beijing PM2.5 and network traffic datasets from UCI~\cite{Dua:2019}. In each case, we used $k = 24$ steps in the past to look ahead, and overall we used the first $30\%$ of the time-steps for training and the rest for testing.

\noindent \textbf{Model Architecture}: Our architecture comprised of a RNN with LSTM units (2 layers with 128 hidden dimensions) and a linear layer for making the final prediction.

\noindent \textbf{Results}: For this study, we ran all experiments with the desired level of calibration fixed at $\alpha = 0.95$, and report the RMSE, $CE_{0.95}$ and $AW_{0.95}$ metrics obtained using the different approaches in Table~\ref{tab:ts}. As it can be seen, the proposed alternating optimization achieves the lowest calibration error in all cases, and more importantly, consistently produces higher-fidelity mean estimators. Figure~\ref{fig:ts} illustrates the prediction intervals obtained using \textit{IQR Fit} for a few cases, where the quality of calibration is clearly evident.

\begin{table}[t]
	\centering
	\renewcommand*{\arraystretch}{1.3}
	\caption{Results from the object localization experiments - We report the performance of the best performing baseline and the proposed \textit{IQR Fit} at $\alpha = 0.95$}
	\begin{tabular}{|c|c|c|c|c|}
		\hline
		& \multicolumn{2}{c|}{ \textbf{HNN} } & \multicolumn{2}{c|}{ \textbf{IQR Fit} } \\ \cline{2-5}
		& $\text{CE}_{0.95}$ & $\text{AW}_{0.95}$ & $\text{CE}_{0.95}$ & $\text{AW}_{0.95}$  \\ \hline
		$x$ & 0.11 & 31.72 & \textbf{0.01} & 18.67 \\
		$y$ & 0.14 & 26.33 & \textbf{0.02} & 16.42 \\
		$w$ & 0.12 & 28.87 & \textbf{0.04} & 20.51 \\
		$h$ & 0.15 & 29.14 & \textbf{0.04} & 21.13 \\
		\hline
		\rowcolor{gray!15} IoU & \multicolumn{2}{c|}{0.468} & \multicolumn{2}{c|}{\textbf{0.557}} \\
		\hline
	\end{tabular}
	\label{tab:obj}
\end{table}

\begin{figure}[t]
	\centering
	{\includegraphics[width=0.14\textwidth]{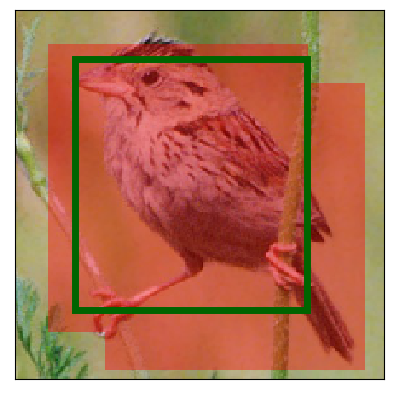}\phantomsubcaption}
	{\includegraphics[width=0.14\textwidth]{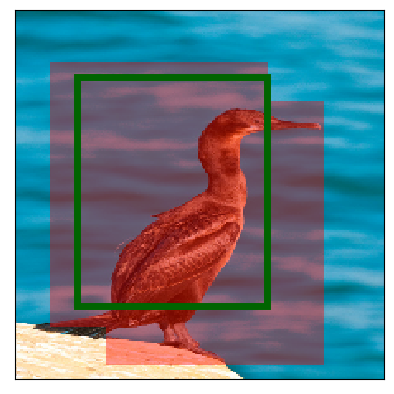}\phantomsubcaption}
	{\includegraphics[width=0.14\textwidth]{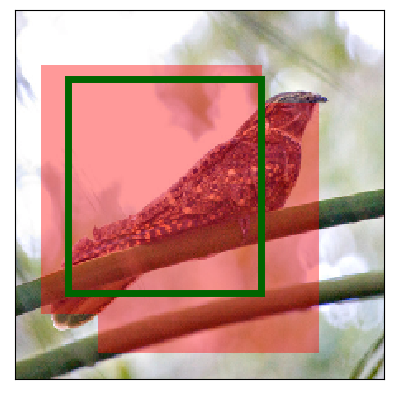}\phantomsubcaption}
	{\includegraphics[width=0.14\textwidth]{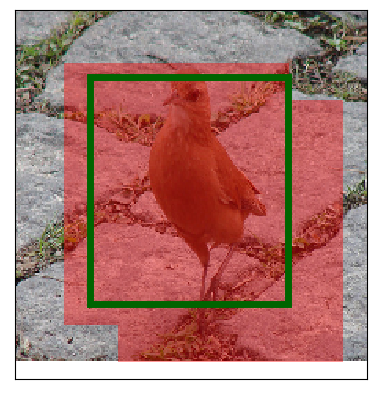}\phantomsubcaption}
	{\includegraphics[width=0.14\textwidth]{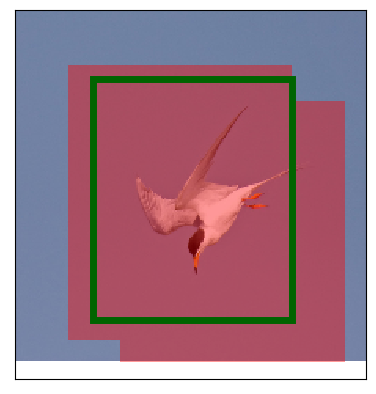}\phantomsubcaption}
	{\includegraphics[width=0.14\textwidth]{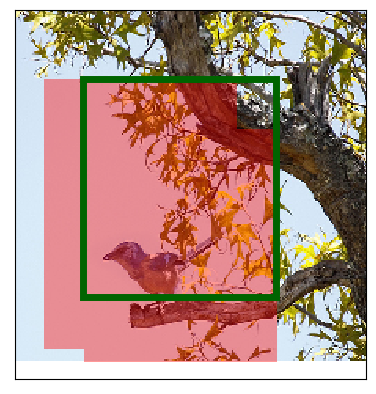}\phantomsubcaption}
	{\includegraphics[width=0.14\textwidth]{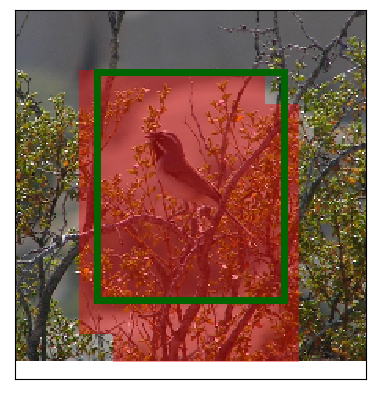}\phantomsubcaption}
	{\includegraphics[width=0.14\textwidth]{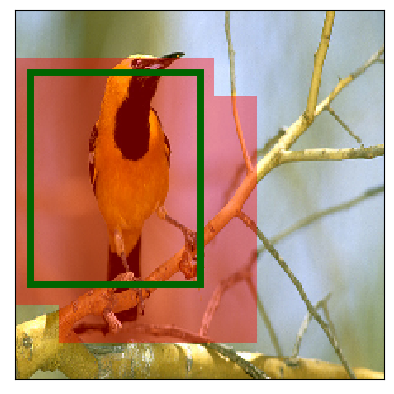}\phantomsubcaption}
	{\includegraphics[width=0.14\textwidth]{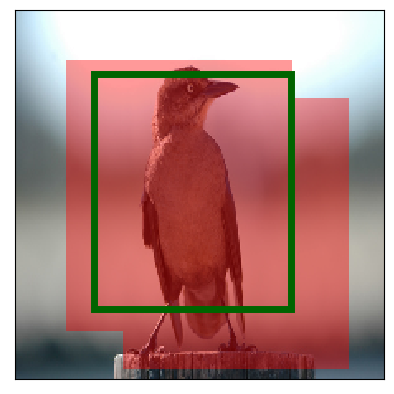}\phantomsubcaption}

	\caption{Illustration of prediction intervals estimated using the proposed \textit{IQR Fit} approach. In each case, the mean prediction is marked in green.}
	\label{fig:obj}
\end{figure}

\subsection{Object Localization using CNNs}
The third study  involved the application of the proposed approaches to object localization in images using convolutional neural networks. We posed the problem of localizing as predicting a bounding box around the object. Each bounding box is characterized by $4$ regression targets -- $(x,y)$ coordinates of the left-bottom corner, height $h$ and width $w$.

\noindent \textbf{Datasets}: For this experiment, we used the Caltech Birds (CUB) dataset~\cite{WelinderEtal2010CUB}, which is comprised of $11,788$ images belonging $200$ different categories of birds.

\noindent \textbf{Model Architecture}: We performed transfer learning from a ResNet-18 model, pre-trained on imagenet classification, to carry out this prediction. We allowed fine-tuning of the last residual block and the final prediction layer during training.

\noindent \textbf{Results}: For brevity, we report results only from the best performing baseline method, HNN, and the proposed \textit{IQR Fit} technique. Similar to the forecasting experiment, we used $\alpha = 0.95$. As seen in Table \ref{tab:obj}, the proposed approach significantly outperforms the baseline in terms of calibration error, while also meeting the sharpness constraint enforced in our optimization. Finally, the IoU score improvement clearly demonstrates the fidelity of the mean estimator. Figure \ref{fig:obj} illustrates the prediction intervals (mean prediction showed in green) obtained using \textit{IQR Fit} for a few example cases.

\section{Conclusions}
In this work, we proposed a framework to build calibrated models for continuous-valued regression problems. By formulating the problems of estimating means and prediction intervals as a bi-level optimization, we showed that the model can effectively capture the informative uncertainties to produce well-calibrated intervals. Further, we found that, this alternating optimization also leads to better regularization of the mean estimator, thus improving its generalization significantly. From our empirical studies, we find that the proposed approach is effective across different data modalities, model architectures and applications. Though our approach was demonstrated using HNN-based and quantile-based aleatoric uncertainty estimators, any existing UQ technique can be integrated into our approach. This work also emphasizes the need to rigorously study the process of producing calibrated intervals based on prediction uncertainties.

\bibliographystyle{aaai}
\bibliography{ref}

\end{document}